%% file: main.tex
\documentclass{article}
\usepackage{booktabs}
\usepackage[ruled,vlined]{algorithm2e}
\usepackage{multirow}
\usepackage{float}
\usepackage{lscape}
\usepackage{stfloats}
\usepackage{xcolor}
\usepackage{bm}
\usepackage{mathtools}
\DeclareMathOperator*{\argmin}{arg\,min}

\definecolor{olivine}{rgb}{0.6, 0.73, 0.45}

\begin{document}

\title{Unsupervised Deep Manifold Attributed Graph Embedding\\}

\author{
Zelin Zang, Siyuan Li, Di Wu, Jianzhu Guo, Yongjie Xu, Stan Z. Li\\
Zhejiang University, Hangzhou, China\\
AI Lab, School of Engineering\\
Westlake University, Hangzhou, China
}

\maketitle

\begin{abstract}
  Unsupervised attributed graph representation learning is challenging since both structural and feature information are required to be represented in the latent space. 
  Existing methods concentrate on learning latent representation via reconstruction tasks, but cannot directly optimize representation and are prone to oversmoothing, thus limiting the applications on downstream tasks.
  To alleviate these issues, we propose a novel graph embedding framework named Deep Manifold Attributed Graph Embedding (DMAGE). A node-to-node geodesic similarity is proposed to compute the inter-node similarity between the data space and the latent space and then use Bergman divergence as loss function to minimize the difference between them. 
  We then design a new network structure with fewer aggregation to alleviate the oversmoothing problem and incorporate graph structure augmentation to improve the representation's stability. Our proposed DMAGE surpasses state-of-the-art methods by a significant margin on three downstream tasks: unsupervised visualization, node clustering, and link prediction across four popular datasets.  
\end{abstract}

\begin{figure}[htb]
  \centering
  \includegraphics[width=4.5in]{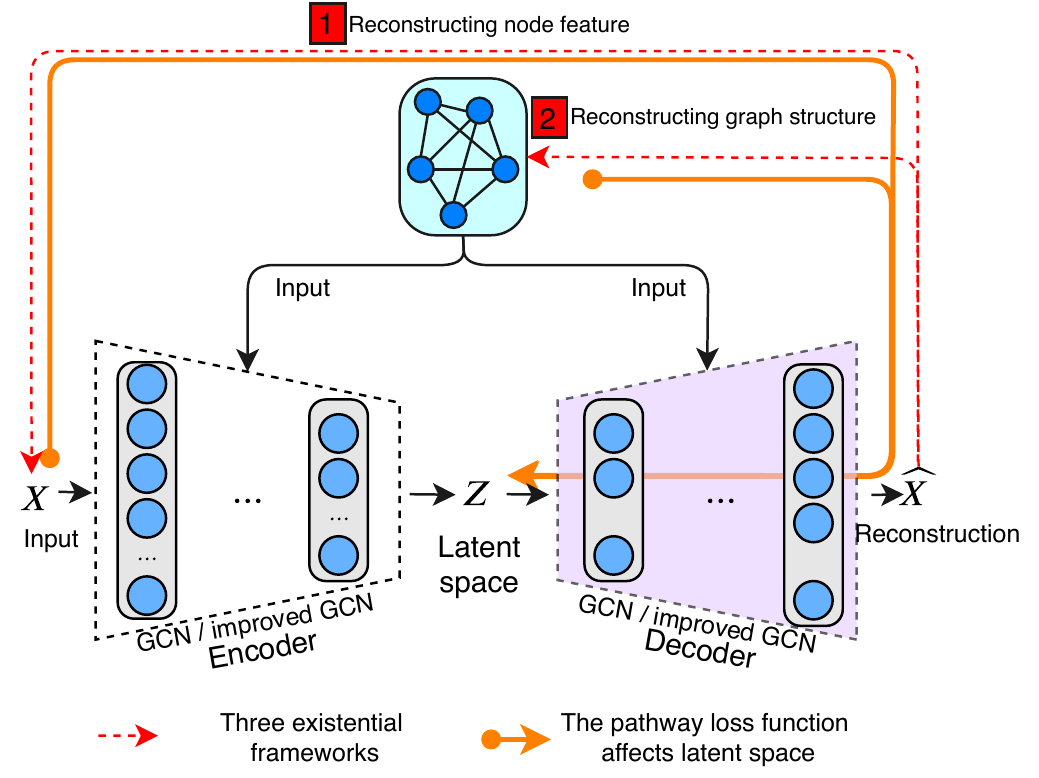}
  \caption{Frameworks of existing neural network-based methods: (1) reconstructing node feature, and (2) reconstructing graph structure. The latent space is optimized through decoders during backpropagation, thus the learned latent representation of all two frameworks are task-dependent and obscured, which leads to the lack of interpretability and subsequent performance guarantees (e.g., generalizability, transferability, and robustness, etc.)}
  \label{fig:intro}
\end{figure}

Attributed graphs are graphs with node attributes/features and are widely applied to represent network-structured data, e.g., social networks \cite{hastings_community_2006}, citation networks \cite{kipf_semi-supervised_2016},
recommendation systems \cite{ying_graph_2018}. Tasks such as analyzing attributed graphs include node clustering, link prediction, and visualization are challenging in high dimensional non-Euclidean space. Graph embedding methods aim to solve these tasks in a low-dimensional latent space that preserves graph information.

Early approaches such as DeepWalk \cite{perozzi_deepwalk_2014}, and GraphSAGE \cite{hamilton_inductive_2018} focus on local information but lack a global view. Autoencoder-based methods such as GAE/VGAE \cite{kipf_variational_2016}, AGC \cite{zhang_attributed_2019}, and DAEGC \cite{wang_attributed_2019} obtain embedding by enforcing the reconstruction constraint. However, the reconstruction task only requires preserving information beneficial of reconstruction and fails to optimize embedding directly (shown in Fig.~\ref{fig:intro}) yet does not guarantee a good representation generalizable to other tasks \cite{cui_adaptive_2020}. Graph Convolutional Network (GCN) based methods, e.g. ARGA \cite{pan_learning_2020}, AGE \cite{cui_adaptive_2020} are usually used to combine graph structure information with node feature information. Such methods suffer from oversmoothing caused by the stacked aggregation layers in the graph network, which loses high-frequency components \cite{wu_simplifying_2019}.

To solve the problems mentioned above, we propose Deep Manifold Attributed Graph Embedding (DMAGE), which is a similarity-based framework for attributed graph embedding, as shown in Fig.~\ref{fig:main}.
Unlike reconstruction based methods, we aim to preserve inter-node similarity between non-Euclidean high dimensional space and Euclidean latent space. Firstly, we propose graph geodesic similarity to collect both graph structure information and node feature information, which measures a node-to-node relationship via the graph's shortest paths. $t$-distribution is used as a kernel function to fit the neighborhoods between nodes instead of Gaussian distribution to balance intra-cluster and inter-cluster relationship. Secondly, to mitigate the oversmoothing phenomenon in the deeply stacked GCN networks, we use a fully connected neural network structure with a less aggregated layer called Fully Connected Aggregation Layer (FCA). The proposed network layout allows a deeper network while taking into account the appropriate aggregation operation. Finally, we design a graph structure augmentation scheme during the training process to improve network embedding stability. We randomly add edges with the hop-1 neighbor and drop edges with the hop-2 neighbors to enhance the graph structure without overly changing the graph semantics, forcing the network to obtain a stable embedding mapping.
Our contributions can be summarized as:
\begin{itemize}
  \item Propose a node-to-node geodesic similarity-based graph embedding framework that preserves both graph structure and node feature information.
  \item Design a network structure to mitigate the oversmoothing problem and a graph structure augmentation method to improve the embedding's stability.
  \item Achieve new state-of-the-art performance on visualization, node clustering, and link prediction tasks on benchmark through extensive experiments.
\end{itemize}

\section{Related Works}

\input{appendix_related_work.tex}

\section{PROPOSED METHOD}

\begin{figure*}[htb]
  \centering
  \includegraphics[width=4.5in]{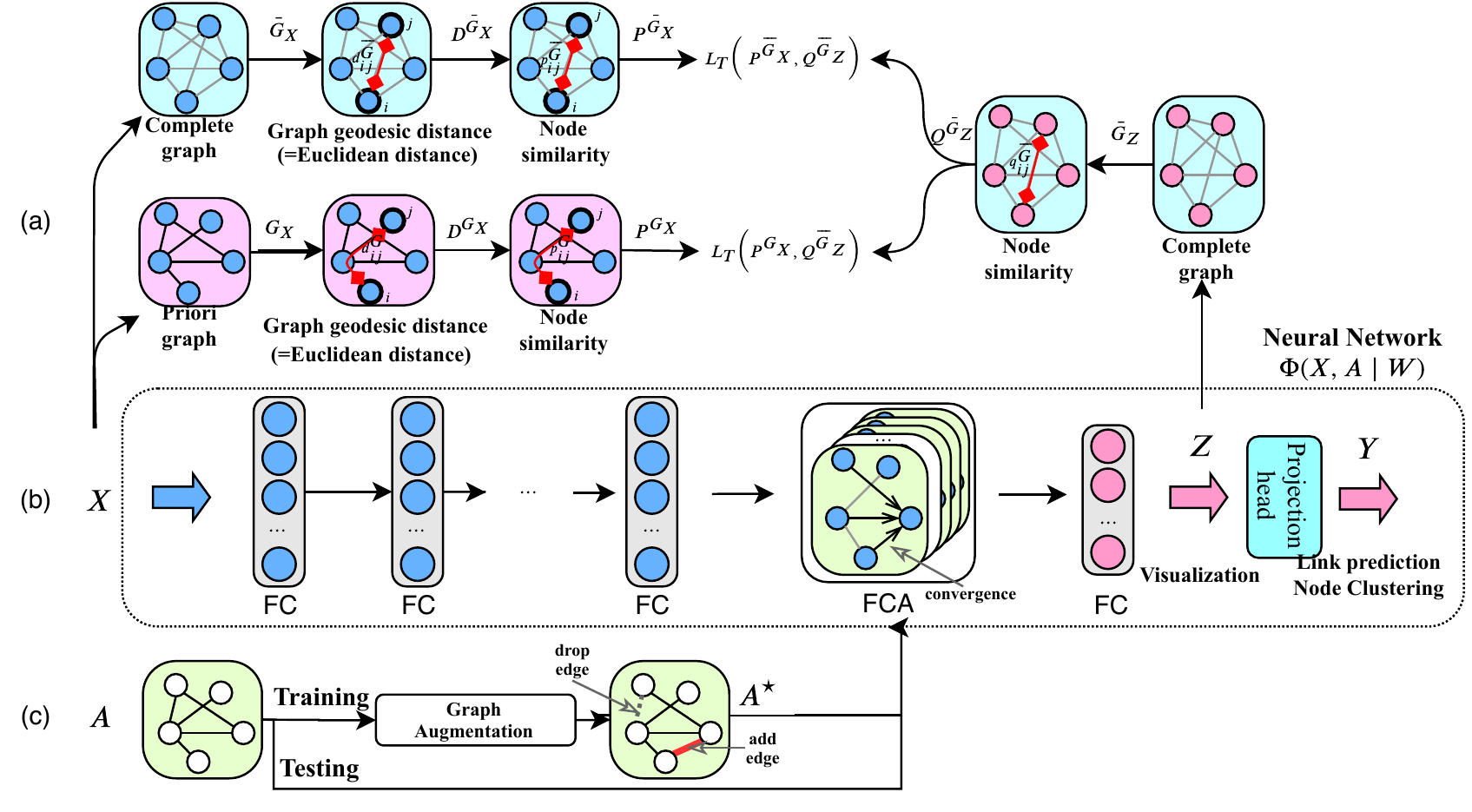}
  \caption{Illustration of Deep Manifold Attributed Graph Embedding (DMAGE) framework: Part (a) describes the fusion process of the structure and feature information in the loss calculation. Graph geodesic distance between nodes is first calculated using input data $X$ and then transformed to similarity on two graphs (Priori graph ${G_X}$ and complete graph $\bar G_X$) correspondingly. We measure the difference of inter-node similarity between the latent space $Z$ and input space. Part (b) describes our proposed network structure and demonstrates the aggregation based on the adjacency matrix in the Fully Connected Aggregation (FCA) layer. Part (c) illustrates the graph augmentation Process. During the training phase, the augmented graph adjacency matrix is passed to FCA, while during the testing process, we directly use the prior graph adjacency matrix.}
  \label{fig:main}
\end{figure*}

In this section, we first formalize the embedding task on attributed graphs. Then we propose our DMAGE algorithm. Specifically, we propose graph geodesic similarity. Then we describe the neural network structure and graph structure augmentation methods. Finally, we define the loss function and algorithmic framework of DMAGE. The overall framework is shown in Fig.~\ref{fig:main}.

\subsection{Problem Formalization}

Given an attributed graph $\bm G=(\bm{V}, \bm{E}, \bm{X})$, where $\bm{V}=\left\{v_{1}, \cdots, v_{n}\right\}$ is the vertex set with $n$ nodes in total, $\bm{E}$ is the edge set, and $\bm{X}=\left[{x}_{1},  \cdots, {x}_{n}\right]^{T}$ is the feature matrix.
The topology structure of graph $G$ can be denoted by adjacency matrix ${\bm A}$:
\begin{equation}
  \bm{A}=\left\{a_{i j}\right\} \in \mathcal{R}^{n \times n}, a_{i j}=1 \text{ if } \left(v_{i}, v_{j}\right) \in \bm{E} \text{ else } 0.
  \label{equ:A}
\end{equation}

We seek to find a set of low-dimensional embeddings $ \bm Z=\left[{z}_{1}, \cdots, {z}_{n}\right]^{T}$ which preserves both structure and feature information of $\bm G$ and use $Z$ to accomplish a variety of downstream tasks like node clustering, link prediction, and visualization.

\subsection{Graph Geodesic Similarity}

{Graph geodesic distance} $\bm D^{\bm G} =\{ d_{ij}^{ \bm G} \ |\  i,j = 1, \cdots, n \}$ describes the shortest distance between nodes on the graph $\bm G[ \bm V,\bm E,\bm X]$, defined as
\begin{equation}
  \begin{aligned}
    d_{ij}^{\bm G} & =  \left\{
    \begin{aligned}
      & \pi(v_i, v_j) \;\;\;\;\;\;\;\;\;\;\;\;  \text{if } v_i, v_j \text{ are connected}     \\
      & \Lambda \cdot \text{max}(\bm{D^G}) \;\;\;\;\;  \text{otherwise}
    \end{aligned}
    \right.
  \end{aligned}
  \label{equ:D_X^G}
\end{equation}
\noindent where $\pi(v_i, v_j)$ is the shortest path based on any distance metric between node $v_i$ and $v_j$. The distance metric can be chosen from various schemes, such as Euclidean distance, Manhattan distance, cosine distance, etc.

We assume that the distance of unconnected nodes is much greater than the maximum distance between other connected nodes. Thus, we denote the distance between unconnected nodes with the maximum distance of connected nodes multiplied by a large positive constant $\Lambda$.

To avoid the adverse effects of outliers and neighborhood inhomogeneity in real data on the characterization of the manifold, we transform the distance $d_{ij}^{\bm G}$  to $d_{i|j}^{\bm G}$ using $\rho_i$ and $\sigma_i^*$.

\begin{equation}
  d_{i|j}^{\bm G} = \frac{{d_{ij}^{\bm G} -\rho_i }}{\sigma_{i}^*},
\end{equation}

\noindent where $\rho_i= \min( [d^G_{i\ 0}, \cdots, d^G_{i\ n }])$ is deducted from distances of all others nodes to node $v_i$ for the purpose of alleviating possible skewed embedding caused by outliers. 
$d_{i|j}^{\bm G}$ and $d_{j|i}^{\bm G}$ have different $\rho_i$ and $\sigma_i$, 
so $d_{i|j}^{\bm G} \neq d_{j|i}^{\bm G}$. 
In addition, the transformation in distance will make the distance between each node and its nearest node to $0$, which means that the similarity is normalized to $1$.
The optimal $\sigma_i^*$ is determined by a binary search method, under the objective function. 

\begin{equation}
  \sigma_i^* = \argmin_{\sigma_i} :|Q_p-2^{\sum_ j \kappa (({{d_{ij}^{\bm G} -\rho_i }})/{\sigma_{i}}  , \nu)^2 }|,
  \label{equ:bin_search}
\end{equation}

\noindent where $Q_p$ and $v$ are hyperparameters that controls the compactness of neighbors, and $\kappa(\cdot)$ is a $t$-distribution kernel function which map the distance $d$ to similarity

\begin{equation}
  \begin{aligned}
    \kappa (d,\nu)  = \sqrt{2\pi} \cdot\frac{\Gamma\left(\frac{\nu+1}{2}\right)}
            {\sqrt{\nu \pi} \Gamma\left(\frac{\nu}{2}\right)}
            \left(
            1+\frac{{d}^{2}}
            {\nu}
            \right)^{-\frac{(\nu+1)}{2}},
  \end{aligned}
  \label{equ:kappa}
\end{equation}

\noindent where $\nu$ is degrees of freedom in $t$-distribution and related discussion is in the ablation study.

We further transform the graph geodesic distance $d^{\bm G}_{i|j}$ to graph geodesic similarity $p^{\bm G}_{i|j}$ by 

\begin{equation}
  p^{\bm G}_{i|j} =  \kappa ( d^{\bm G}_{i|j},\nu).
  \label{equ:P_ij^G}
\end{equation}

Then, {\bf Graph geodesic similarity} $p^{\bm G}_{i|j}$ is symmetrized in the form of joint probability using the following equation:
\begin{equation}
  p^{\bm G}_{ij} = p_{i|j}^{\bm G} + { p_{j|i}^{\bm G} } - 2 p_{i|j}^{\bm G} { p_{j|i}^{\bm G} }.
  \label{equ:simi}
\end{equation}
In summary, we write the calculation of graph geodesic similarity in matrix form.
\begin{equation}
  \bm P^{\bm G} = S(\bm G[\bm V,\bm E,\bm X], \nu)=\{p_{ij}^{\bm G}| i,j=1,\cdots,n\}.
  \label{equ:ggdall}
\end{equation}

\subsection{Network Structure and Graph Augmentation}
Excessive aggregation operations and undesired filtering in traditional GNNs bring about the problem of oversmoothing. We use the conventional fully connected layer (FC) and fully connected aggregation layer (FCA) to assemble our neural network to combat the oversmoothing problem. Also, the structure we apply allows for a deeper network and, therefore, better mapping capabilities.

{\bf Fully Connected Aggregation layer (FCA)}:

\begin{equation}
  \begin{aligned}
    \!\bm Z^{\!l+1} \!&= \text{\!FCA}^l\!( \bm Z^{l}\!,\bm A| \bm W^{l}\!,\bm B^{l}\!) \\
    &=\sum_{(i,j)\in \bm{E}} \sqrt{|\mathcal{N}(i)|} \sqrt{|\mathcal{N}(j)|} \  (w^{l}_{ij} z^{l}_j + b^l_j).
\end{aligned}
  \label{equ:FCA_M}
\end{equation}

\noindent where $|\mathcal{N}(i)|$ is number of neighbors of node $v_i$, $\bm Z^{l}$ being the feature matrix of layer $l$. $\bm W^{l}$, $\bm B^{l}$ is the weight and bias of layer $l$ respectively.
FCA can be viewed as a GCN \cite{kipf_semi-supervised_2016} layer without activation function.

A deeper network structure with $L$ layers $\Phi(\bm X, \bm A| \bm{W}, \bm{B})$ with fewer aggregation operations can be defined as

\begin{equation}
  \begin{aligned}
    \bm Z^{1}   & = \text{FC}^1(\bm X | \bm W^{1}, \bm B^{1})               \\
    \bm Z^{2} & = \text{FC}^{2}(\bm Z^{1} | \bm W^{2}, \bm B^{2})     \\
            & \cdots                                            \\
    \bm Z^{L-1} & = \text{FCA}^{L-1}(\bm Z^{L-2},\bm A | \bm W^{L-1}, \bm B^{L-1}) \\
    \bm Z       & = \text{FC}^L( \bm Z^{L-1} | \bm W^{L}, \bm B^{L}) .      \\
  \end{aligned}
  \label{equ:networt_layers}
\end{equation}

In general, the forward propagation of our model can be summarized as

\begin{equation}
  Z = \Phi(X, A| \textbf{W} ),           
  \label{equ:phi}
\end{equation}

In complex networks, such as social networks, the relationship of nodes is often considered stable. A graph's semantic information does not change dramatically due to changes in particular edges. A useful graph embedding should also behave durable against edge changes. Therefore, we introduce a graph augmentation method during the training process to improve the embedding's stability and reduce the dependence of the model on the correctness of the graph structure.
For descriptive convenience, the enhancemented edge $\bm E^\star$ is defined as (can be easily transformed into an adjacency matrix $\bm A^*$ by Eq.~(\ref{equ:A})) 

\begin{equation}
  \begin{aligned}
    & \bm E^\star  = \bm E - \bm E^{-} + \bm E^{+} 
  \end{aligned}
\end{equation}

\noindent where $\bm E^{-}$ is a set of edges to be removed, $\bm E^{+}$ is a set of edges to be added.

\begin{equation}
  \begin{aligned}
    & \bm E^{-} = \{ \mathbf{1}_{r_{ij} > p^- } (i, j) \ \  |\  , i \in V, j \in \bm H_1(i) \} \\
    & \bm E^{+} = \{ \mathbf{1}_{ r_{ij} > p^+ } (i, j) \ \  |\  , i \in V, j \in \bm H_2(i) \} \\
    & r_{ij} \in \bm U(0,1),
  \end{aligned}
    \label{equ:hatE}
\end{equation}

\noindent where $\bm H_1(i)$ and $\bm H_2(i)$ are the hop-1 neighborhood and hop-2 neighborhood of node $i$. $p^{-}$ and $p^{+}$ are the probability of augmentation. In this paper, the number of added and removed edges is kept equal in our experiments. During the training process, edge augmentation is performed randomly at the beginning of each epoch under uniform distribution $\bm U(0,1)$.

\subsection{Bregman divergence and Information Fusion}
Based on the graph geodesic similarity, this paper uses Bregman divergence as loss function to minimize the similarity difference between two spaces. Bregman divergence is defined as follows:

\begin{equation}
  \begin{aligned}
    L_T\!(x,y,\ | \ F(\cdot) )\!= 
    \{ 
       F(x)\!-\!F(y)\! 
       -\!\langle\nabla F(x), x\!-\!y\rangle 
    \}.
  \end{aligned}
  \end{equation}

\noindent where $F(\cdot)$ is any continuously differentiable, strictly convex function defined on a closed convex set. Our model minimizes the difference in node similarity between the input space and the latent space. We find that it is unnecessary to carefully choose the metric difference ground loss function $F(\cdot)$. We give the following examples of Bregman divergence with different choices. 

\begin{itemize}
  \item Squared Euclidean Distance (DMAGE-SED for short)
 \begin{equation}
  L_T\!(\bm x,\bm y\ \left.\right| \ {\|\bm x\|^2})\!= \|\bm x-\bm y\|_2 .
  \label{equ:L^GSED}
\end{equation}
  \item Logistic Loss (DMAGE-LOGI for short)
 \begin{equation}
  \begin{aligned}
    L_T\!(\bm x,&\bm y\ \left.\right|\ \bm x\log(\bm x)\!+\!(1\!-\bm x)\!\log(1\!-\bm x\!) )\!=\\ 
            &
            \{
              \bm x\log \frac{\bm x}{\bm y} 
              \!+\!(1-\bm x\!)\log \frac{1-\bm x}{1-\bm y}
            \}.
  \end{aligned}
  \label{equ:L^GLOGI}
\end{equation}
\end{itemize}

Several decisions of $F(\cdot)$ are compared in the experiment section.

Finally, we complete the definition of overall loss function defined with the fusion of graph structure information and feature information as follows: 

\begin{equation}
  \begin{aligned}
  L &= L_T(  \bm{P^{\overline G_X}}, \bm{P^{\overline G_Z}} ) + \alpha \cdot L_T( \bm{P^{G_X}}, \bm{P^{\overline G_Z}} ),\\
  \end{aligned}
  \label{equ:LmathcalG}
\end{equation}

\noindent where  $\alpha$ is a balancing parameter between two terms. A bar on top of a graph $\bm G[\bm{V},{\bm{E}},\bm X]$ (short for $\bm{G}$) means the new graph $\bm{\overline G}[\bm{V},\overline{\bm{E}},\bm X]$ (short for $\bm{\overline G}$) is a complete graph with same nodes defined in $\bm G$. $\bm{P^{\overline G_X}}$, $\bm{P^{G_X}}$ and $\bm{P^{\overline G_Z}}$ is geodisic similarity defined on each corresponding graph using Eq.~(\ref{equ:ggdall}):

\begin{equation}
  \begin{aligned}
  \bm{P^{G_X}}             &= S( \bm{G_X}           (\bm{V},\bm E,\bm X), \nu_{input} ),\\
  \bm{P^{\overline{G_{X}}}}&= S(\overline{\bm{G_X}} (\bm{V},\overline{\bm{E}},\bm X), \nu_{input} ),\\
  \bm{P^{\overline{G_Z}}}  &= S(\overline{\bm{G_Z}} (\bm{V},\overline{\bm E},\bm Z), \nu_{latent} ).\\
  \end{aligned}
  \label{equ:cP}
\end{equation}


Term $L_T(\bm{P^{\overline G_X}}, \bm{P^{\overline G_Z}} )$ computes inter-node similarity on the complete graph ignoring any graph structure information. It ensures node feature information is preserved in the embedding. Term $L_T(\bm{ P^{G_X}}, \bm{Q^{\overline G_Z}} )$  takes graph structure into account while minimizing similarity difference. Combining two loss terms guarantees the latent embedding captures both feature and structure information. We use different $t$-distribution degrees of freedom coefficients $v_{input}$ and $v_{latent}$ to deal with the crowding problem\cite{maaten_visualizing_2008} caused by the different dimensions of the two spaces, as we will illustrate in the ablation experiments. Also, this loss function design with multiple information fusion can be extended to multi-graph node embedding.

\subsection{Algorithm and Complexity}

\input{main_pseudoCode.tex}

Algorithm.~\ref{alg:DMAGE} shows how to train a DMAGE model and obtain the attributed graph representation. 
The overall complexity is lower than $O(|V|^2)$, where $|V|$ is the number of edges. We divide the algorithm into two parts: the initialization part and the network training part. 
The initialization part calculates the graph geodesic similarity. It costs $O(|V|^2)$ to computie the pair-wise distance matrix of a fully connected graph. To reduce the complexity, we use a $K$-nearest neighbor graph instead of the fully connected graph. Thus, the K-nearest neighbor graph's time complexity is determined as $O(K|V|^{1.14})$ via the Nearest-Neighbor-Descent algorithm \cite{kobak_umap_2019}.
The traditional Dijkstra algorithm to solve the shortest path, the time complexity is $O(|E|+|V| \log |V|)$, where $|E|$ is the number of nodes.
The network training part uses a well-established mini-batch-based training method, and the complexity is $O(B_s|V|)$, where $B_s$ is the batch size.

\section{EVALUATION ON PUBLIC DATASETS}

In this section, we evaluate the effectiveness of DMAGE with the state-of-the-art methods on different unsupervised attributed graph representation tasks. Then, we explore the effectiveness of the proposed modules and robustness of hyperparameters.


\subsection{Datasets and Experimental Setup}

\textbf{Datasets.} We conduct experiments on four widely used network datasets (CORA, CiteSeer, PubMed \cite{kipf_semi-supervised_2016} and Wiki \cite{yang_network_2015}). Features in CORA and CiteSeer are binary word vectors, while in Wiki and PubMed, nodes are associated with tf-idf weighted word vectors. The statistics of the four datasets are shown in table \ref{Tab_info_4dataset}.  

\input{Tab_info_4dataset.tex}

\noindent \textbf{Experimental Setup.} For all datasets, 
the neural network structure is set to be [FC(input dimension) $\to$ FC(500) $\to$ FC(250) $\to$ FCA(250) $\to$ FC(200)]. 
The latent space's dimension is 200. 
The $t$-distribution degrees of freedom of the input data are set to $v_{input}=100$. 
The dropding edge rate $p^-=0.01$, the adding edge rate $p^+=|E^-|/|H_2|$, where $|H_2|$ is the number of the edges that have the potential to be added. 
$\Lambda=10$. 
Cosine distance is used in Eq.~(\ref{equ:D_X^G}). 
We directly set $\sigma_i^*=1$ and $\rho_i=0$ in the latent space as a trade-off between the speed and performance. 
Only the latent space degrees of freedom $v_{latent}$, weights $\alpha$ and $Q_p$ are selected accordingly for each dataset, details are shown in Appendix.
All codes are implementated using PyTorch library and run on NVIDIA v100 GPU.

\subsection{Baseline Methods}

\input{Tab_result_node_clustering.tex}

We compare our model with eight more algorithms. The baselines can be categorized into three groups:

(1) \textbf{Methods using features only}. k-means [20] and Spectral Clustering (Spectral-f) are two traditional clustering algorithms. Spectral-f takes the cosine similarity of node features as input.

(2) \textbf{Methods using graph structure only}. Spectral-g is Spectral Clustering with the adjacency matrix as the input similarity matrix. DeepWalk \cite{perozzi_deepwalk_2014} learns node embeddings using SkipGram on generated random walk paths on graphs.

(3) \textbf{Methods using both features and graph}. We compared some classical algorithms such as GAE,  VGAE \cite{kipf_variational_2016}, MGAE \cite{wang_mgae_2017},  ARGE \cite{pan_adversarially_2018}, DANE \cite{gao_deep_2018}, DGI \cite{velickovic_deep_2018}, AGC \cite{zhang_attributed_2019} and DAEGC \cite{wang_attributed_2019}, ARGA \cite{pan_learning_2020}. 
And some emerging methods: AGE \cite{cui_adaptive_2020} design a non-parametric Laplacian smoothing filter which preserves optimal denoising properties to filter out high-frequency noises. GIC \cite{mavromatis_graph_2020} (Concurrent Work) learn a graph representation by maximizes the mutual information between similar nodes.

\subsection{Task of Node Clustering}

In the node clustering task, the computed embeddings are clustered into K = \#classes clusters with K-means in an unsupervised manner. Then we evaluate clustering performance using external labels. Also, we use three specific choices of $F(\cdot)$ in Bregman divergence in DMAGE: DMAGE-SED, DMAGE-SED, and DMAGE-SED+LOGI (DMAGE-SED+LOGI is the combination of DMAGE-SED and DMAGE-LOGI).
We report classification accuracy (ACC), normalized mutual information (NMI), and balanced F-score (F1) in Table \ref{Tab_result_node_clustering}. The best results for each indicator are shown in bold. For the method with the same testing protocol, we directly use the results reported in their paper. For a fair comparison, we used k-means instead of AGE's spectral clustering in the official code to re-run the test. Table \ref{Tab_result_node_clustering} reports the results obtained when the random seed is set to 1. The average results for the 20 runs are shown in Appendix.

DMAGE outperformed state-of-the-art methods in 11 of all 12 tests on PubMed, CORA, and Kiwi. In particular, performance is improved by over 6\% on average on the relatively large PubMed dataset. On Citeseer, DMAGE yields better performance in Acc and F1 scores with a slightly lower NMI result. It's suggested that by preserving similarity information in the embedding space, DMAGE has the advantage in clustering nodes over other reconstruction and adversarial based approaches. 
The performance of our proposed three loss functions is similar.



\subsection{Task of Link Prediction}

\input{Tab_result_linkprediction.tex}

In the link prediction task, some edges are hidden randomly in the input graph, and the goal is to predict the existence of hidden edges based using the computed embeddings. 
We follow the setup described in \cite{kipf_variational_2016,pan_adversarially_2018}: We use 5\% of edges and negative edges as validation set, 10\% of edges and negative edges as the test set. Results are averaged over 20 runs and are shown in table \ref{Tab_result_linkprediction}.
We report the area under the ROC curve (AUC) score \cite{bradley_use_1997}, which equals the probability that a randomly chosen edge is ranked higher than a randomly chosen negative edge. The average precision (AP) score \cite{su_relationship_2015}, which is the area under the precision-recall curve.
We use the same hyperparameters selected for the node clustering as a demonstration of robustness towards hyperparameters.

As shown in table \ref{Tab_result_linkprediction}, our method achieves the highest average value in CORA and CiteSeer data with relatively high stability. VGAE and ARGA perform slightly better than our DMAGE on the PubMed dataset. We discover that VGAE and ARGA are both autoencoder based method. We suspect that embedding acquired with reconstruction supervision preserves more information of the input space, which helps predict hidden edges.

We also find that DMAGE yields similar results for node clustering and link prediction tasks using three specific Bergman divergence as loss functions (DMAGE-SED: squared Euclidean distance, DMAGE-LOGI: logistic loss, DMAGE-SED+LOGI: their combination). It shows that DMAGE is robust to different $F(\cdot)$ in Bregman divergence. Three different Bregman divergence loss functions have similar performance, so we choose only one loss function (DMAGE-LOGI) for visualization, ablation study, and parameter sensitivity analysis.

\subsection{Task of Unsupervised Visualization}

\begin{figure*}[htb]
  \centering
  \includegraphics[width=4.5in]{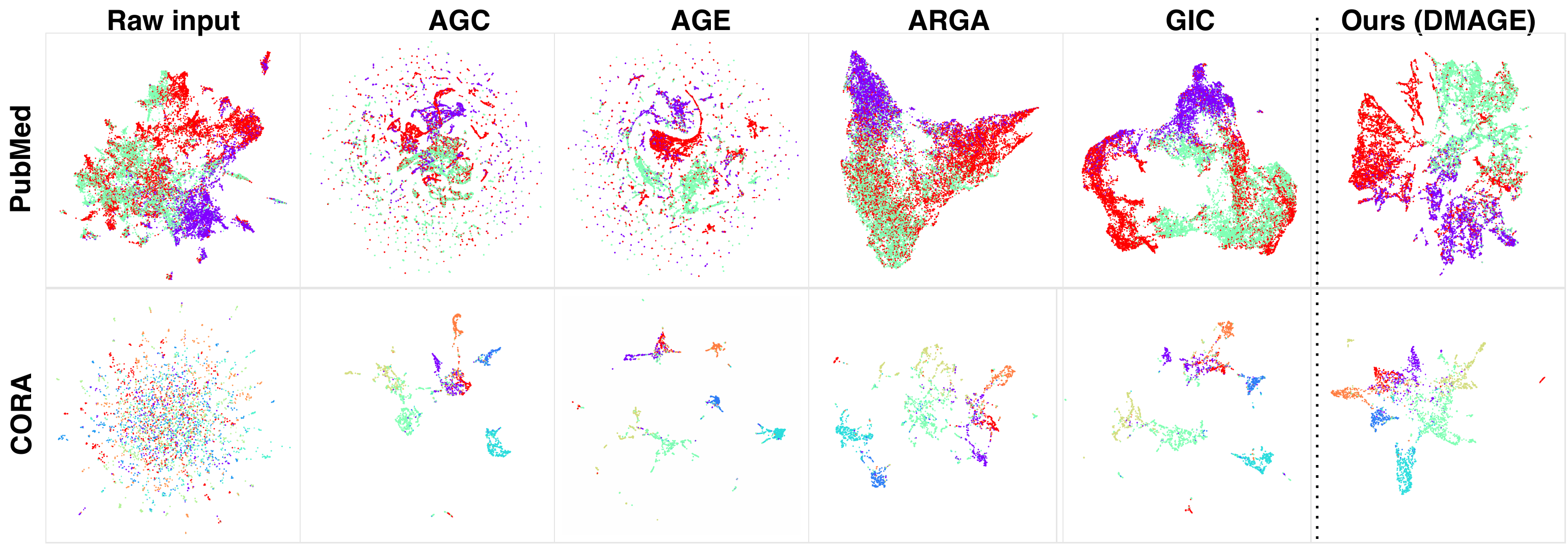}
  \caption{Comparison of visualization results generated by DMT and other methods. Colors represent label categories. Our embedding has clear class boundaries and still preserve the relationship between clusters, which may be a reason for the advantage in both node clustering and link prediction tasks.}
  \label{fig:Fig_vis_main}
\end{figure*}

To evaluate our proposed method, we visualize the distribution of learned latent representations compared to each node's input features in two-dimensional space using UMAP \cite{mcinnes_umap_2018}. We show our method's visualization result compared with other methods on the PubMed and CORA in Fig.~\ref{fig:Fig_vis_main}. We generated the embedding results of the baseline method with the official code. The visualization results show that the DMAGE method can produce more clear category boundaries while preserving the interrelationships between clusters. Visualization results of the all datasets are shown in Appendix.

\subsection{Ablation Study}

To analyze the effectiveness of each module in the proposed methods, we remove them one by one on the node clustering task in Table \ref{Tab_result_ablation}. 

(W/o augmentation): Removing the graph augmentation in the training process and directly using the graph structure given in the aggregation operation dataset. Performance degradation was observed for all datasets. Performance degradation is significant for PubMed and Wiki.

(W/o FCA): We observed a performance decrease on most datasets after removing the FCA layer and replacing it with a regular FC layer. In addition, if all five neural network layers are replaced with GCN layers, the embedding result collapses due to oversmoothing.

(W/o ``soft” similarity): Removing graph geodesic similarity and using "hard" connection relations (1 if there is an edge between two nodes, 0 otherwise).

\begin{figure*}[htb]
  \centering
  \includegraphics[width=4.5in]{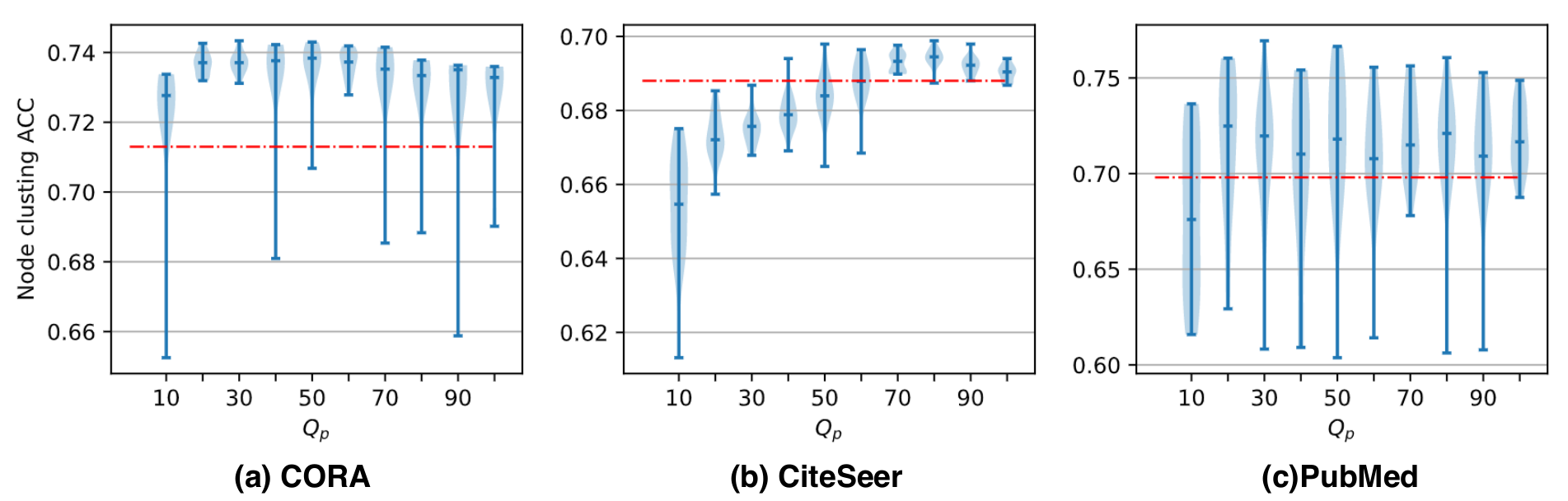}
  \caption{Violin diagram of the node clustering ACC (accuracy) by changing $Q_p$.  The points inside each bar are the median of 20 replicate experiments. When submitting this paper, the highest ACC reported by other methods are marked with the red line (exclude concurrent work).}
  \label{fig:Q_latent}
\end{figure*}

\input{Tab_result_ablation.tex}

All three innovations result in performance improvements, among which graph geodesic similarity brings about most performance gain. The similarity calculation method allows points that are not connected by edges to be related through the graph's node similarity measurement. The experiments also show that the graph node embedding framework based on the inter-node similarity proposed in this paper can improve performance.

\subsection{Parameter Sensitivity Analysis}

We illustrate the mechanism of the method and the robustness of the method by varying the hyperparameters in DMAGE.

\begin{figure}[!]
  \centering
  \includegraphics[width=3.5in]{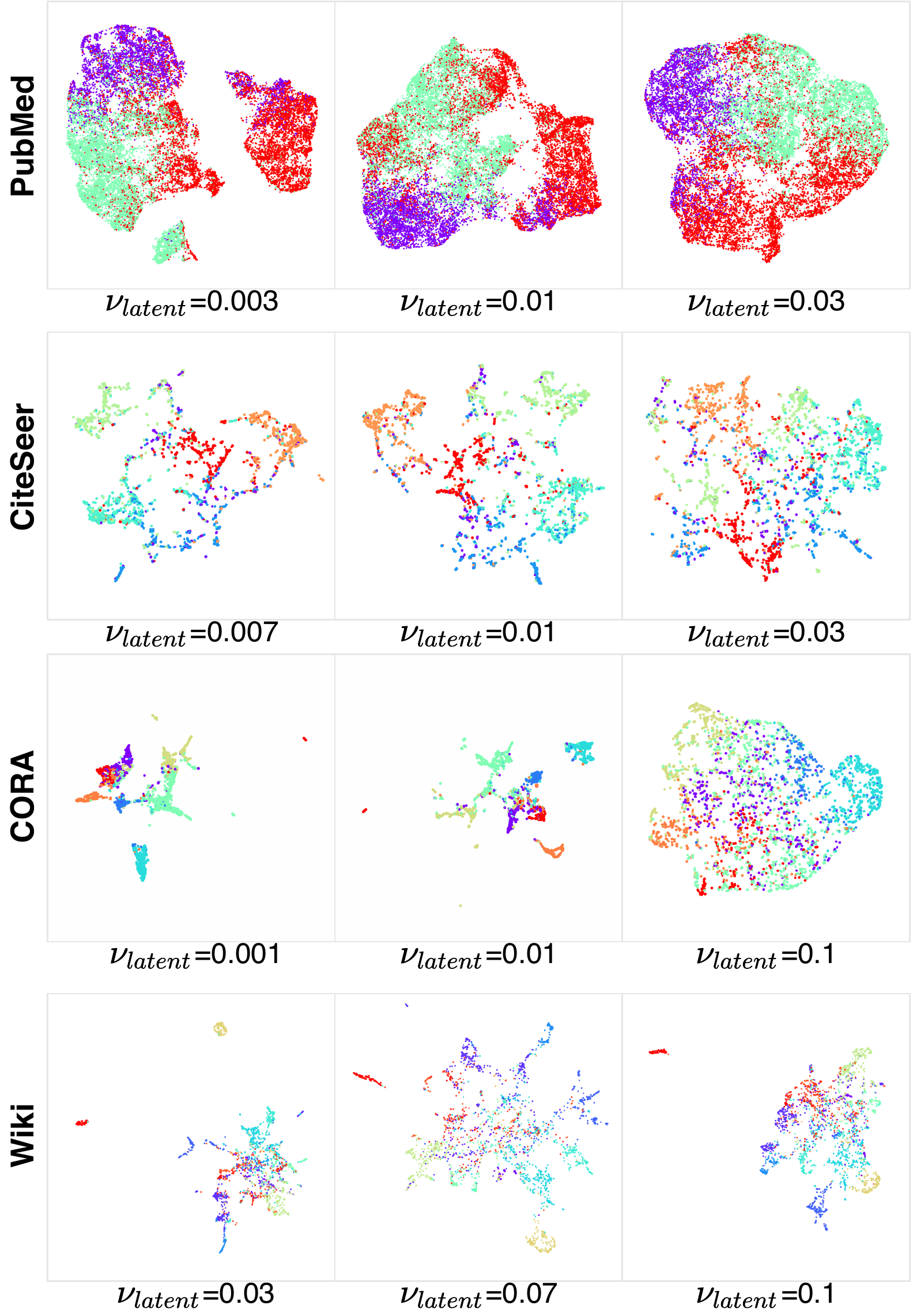}
  \caption{Visualization results of different $\nu_{latent}$. When $\nu_{latent}$ is large, the nodes are close to each other, and the clusters will tend to be distributed together. When $\nu_{latent}$ is small, nodes that are not similar will be pushed further away, and clusters will tend to be separated from each other.}
  \label{fig:nu_latent}
\end{figure}

{\bf Effect of $\nu_{latent}$.} The degree of freedom $\nu$ of the $t$-distribution controls the sharpness of the kernel function. When high-dimensional data is mapped into a low-dimensional space, a crowding problem\cite{maaten_visualizing_2008} may occur. It is essentially due to the lack of expressiveness of the low-dimensional space. The advantage of using the $t$-distribution as a kernel function is that when a crowding problem is encountered, we can change the shape of the mapping function by increasing the latent space degrees of freedom $\nu_{latent}$ (in latent space, this reduces the similarity of neighboring nodes and increases the similarity of non-neighboring nodes), eventually pushing the node's neighbors farther away and alleviating the crowding problem. To illustrate the above idea, we show the visualization results for four datasets with different $\nu_{latent}$ in Fig.~\ref{fig:nu_latent}, and more comments are provided in the accompanying notes below the figure.

{\bf Effect of $Q_p$.} 
The hyperparameter $Q_p$ can help DMT to make a sufficient balance between local and global structure preservation. When $Q_p$ is small, a relatively small $\sigma$ is obtained by binary search, which results in a smaller graph geodesic similarity $p^{\bm G}_{ij}$ will show more local information. As $Q_p$ increases, the $\sigma$ obtained becomes larger, so $p^{\bm G}_{ij}$ becomes larger, and the embedding will present structural information more globally. Fig.~\ref{fig:Q_latent} shows the effect of $Q_p$ on the node clustering average ACC over 20 experiments on three datasets (CORA, CiteSeer, and PubMed). The $Q_p$ can effectively affect the performance of the clustering task but is not very sensitive. Experiments show that we can find an interval of $Q_p$ that allows the DMAGE's average performance to exceed baseline method.

\section{Conclusions and further work}
We proposed an attributed graph node embedding framework based on the node-to-node similarity. We offer an inter-node similarity ground measure and use Bergman divergence as a loss function to optimize the latent space ground embedding. Also, we propose a depth model with less aggregation to avoid oversmoothing and use graph augmentation methods to improve the embedding's stability. We studied our designed components and showed their effectiveness. The experiments proved the method's effectiveness and stability and demonstrated that our innovative parts could substantially lead to enhancements. 

In the future, we expect to extend the method to semi-supervised and supervised domains and to be able to solve some more challenging tasks, such as label propagation and label noise.

\clearpage
\bibliographystyle{plain} 
\bibliography{GELIS}

\clearpage

\appendix
\onecolumn

\input{appendix_nodeclustingmutirun.tex}
\section{Experimental Setup details}
For all datasets, 
the neural network structure is set to be [FC(input dimension) $\to$ FC(500) $\to$ FC(250) $\to$ FCA(250) $\to$ FC(200)]. 
The latent space's dimension is 200. 
The $t$-distribution degrees of freedom of the input data are set to $v_{input}=100$. 
The dropding edge rate $p^-=0.01$, the adding edge rate $p^+=|E^-|/|H_2|$, where $|H_2|$ is the set of all edges that have the potential to be added. 
$\Lambda=10$. 
Cosine distance is used in Eq.~(\ref{equ:D_X^G}). 
To increase the computational speed, we directly set $\sigma_i^*=1$ and $\rho_i=0$ in the latent space. All codes are implementated using PyTorch library and run on NVIDIA v100 GPU. PubMed and Wiki's $\alpha$ are large in alpha parameters because their nodes are associated with tf-idf weighted word vectors

\input{Tab_param.tex}

\section{Comparison of visualization of all datasets}

To evaluate our proposed method, we visualize the distribution of learned latent representations compared to each node's input features in two-dimensional space using UMAP \cite{mcinnes_umap_2018}. We show our method's visualization result compared with other methods on the all datasets in Fig.~\ref{fig:vis1_appendix}. We generated the embedding results of the baseline method with the official code. The visualization results show that the DMAGE method can produce more exact category boundaries while preserving the interrelationships between clusters.

\input{appendix_visualization.tex}

\end{document}

%% file: appendix_related_work.tex
Attributed graph embedding, also known as network embedding \cite{cai_comprehensive_2018} or  network representation learning \cite{zhang_network_2020}, transfers graph nodes into vectors. From the perspective of information exploration, algorithms can be divided into model-free and model-based methods.

We summarize {\bf model-free} methods into four types:
(1) Laplacian eigenmaps-based models \cite{newman_finding_2006},
(2) Local similarity-based models,
(3) Matrix factorization-based models,
and (4) Nonparametric Bayesian modeling-based models.
The local similarity-based approach is based on local similarity with SkipGram model \cite{le_distributed_2014} for embedding, which can handle very large datasets, mainly including
\cite{perozzi_deepwalk_2014,tang_line_2015,grover_node2vec_2016}. 
\cite{li_community_2018,wang_semantic_2016,yang_network_2015} 
is matrix factorization extension that add feature-related regularization terms.
\cite{bojchevski_bayesian_2018,chang_relational_2009} 
model features as latent variables in Nonparametric Bayesian networks.

And the {\bf model-based} methods can be divided into two parts according to whether they use graph convolutional networks (GCN) framework \cite{kipf_semi-supervised_2016} or not.

As a generalization of convolutional operations on graphs, there has been a surge of research interests in graph neural networks in recent years.

(1) Some of these approaches use GCNs as network components and rely on graph autoencoders to fuse graph structure information with feature information.
For examples, graph autoencoder (GAE) and variational graph autoencoder (VGAE) \cite{kipf_variational_2016} learn the node embeddings by using GCN as the encoder, then decode by inner product with cross-entropy loss.
As variants of GAE (VGAE),
\cite{pan_adversarially_2018} exploits adversarially regularized method to learn more robust node embeddings. 
\cite{wang_attributed_2019} further employs graph attention networks \cite{velickovic_graph_2017} to differentiate the importance of the neighboring nodes to a target node.

(2) Some other methods use adjacency matrix as a filter to fuse graph structure information and feature information during forward-propagation of the network.
\cite{wang_mgae_2017} leverages marginalized denoising autoencoder to disturb the structure information.
To build a symmetric graph autoencoder, \cite{park_symmetric_2019} proposes Laplacian sharpening as the counterpart of Laplacian smoothing in the encoder. The authors claim that Laplacian sharpening is a process that makes the reconstructed feature of each node away from the centroid of its neighbors to avoid over-smoothing. 
\cite{zhang_attributed_2019} propose an adaptive graph convolution method for attributed graph clustering that exploits high-order graph convolution to capture global cluster structure and adaptively selects the appropriate order for different graphs.
A better Adaptive Graph Encoder is designed in \cite{cui_adaptive_2020} to smooth and alleviate the high-frequency noises and iteratively strengthen the filtered features. 
\cite{zhu_cagnn_2020} refine the graph topology by strengthening intra-class edges and reducing node connections between different classes based on cluster labels, which better preserves cluster structures in the embedding space.

Also, many schemes for graph node embedding without using GCN networks have emerged due to exploring the GCN over-smoothing problem. 
\cite{gao_deep_2018} uses two networks to learn structural information and feature information separately and then fuses the two kinds of information in latent space. 
\cite{tu_structural_2018} uses a neural network architecture to fuse first-order proximity and second-order proximity of hypergraph networks to complete the node embedding of hypergraph networks. 
\cite{pan_learning_2020} introduced the idea of generative adversarial networks to the graph embedding problem and designed an adversarially regularized graph autoencoder to improve the robustness of embedding.
In addition, \cite{tu_deep_2018,ji_smoothness_2020} use the structure of RNN for node embedding related studies.

%% file: main_pseudoCode.tex
{
    \begin{algorithm}[ht]
        \caption{The DMAGE algorithm}
        \SetAlgoLined
        \textbf{ Input }: Graph: $G[{V}, {E}, {X}]$, \\
        {
            Learning rate: $l_r$, 
            Epochs: $E_p$,
            Batch szie: $B_s$,\\
            Network structure: $N_S$,
            Weight hyperparameter: $\alpha$,\\
            Degree of freedom: $\nu_{latent}$, Graph augmentation rate: $p^+,p^-$\\
            Neighbor prior distribution parameter: $Q_p$ \\
            \textbf{ Output }: 
            Graph Embedding: $Z$, 
            \ \\
            \textcolor{green!55!blue}{\# Initialization}\\ 
            Calculate $d_{ij}^{G_X}$ and $d_{ij}^{\overline G_X}$ by Eq.~(\ref{equ:D_X^G})  \\
            Calculate $\sigma^*$ by Eq.~(\ref{equ:bin_search}) \\
            Calculate $p^{G_X}_{i|j}$ and $p^{\overline G_X}_{i|j}$ by Eq.~(\ref{equ:kappa}) - Eq.~(\ref{equ:P_ij^G})\\
            Calculate $P^{G_X}$ and $P^{\overline G_X}$ by Eq.~(\ref{equ:ggdall})\\
            Initialize the network $\phi(\ \ \cdot\ \ |W)$ with $N_S$\\
            \textcolor{green!55!blue}{\# epcoh loop}\\ 
            \While{$i=0$; $i<E_p$; $i$++}{
                \textcolor{green!55!blue}{\# batch loop}\\ 
                \While{$b=0$; $b<[ |V| /B_s]$; $b$++}{
                Graph structure augmentation by Eq.~(\ref{equ:hatE}), $A^*\leftarrow A_{ug}(A, p^+, p^-)$\\
                Network forward propagation by Eq.~(\ref{equ:networt_layers}), $Z \leftarrow \phi(X, A^*|W)$\\
                Calculate $D^{\overline G_Z}$ by Eq.~(\ref{equ:D_X^G}).\\
                Calculate $Q^{\overline G_Z}$ by Eq.~(\ref{equ:P_ij^G}) and (\ref{equ:simi})\\
                Calculate the loss by Eq.~(\ref{equ:LmathcalG})
                $ L\leftarrow L_T(  P^{\overline G_X}, Q^{\overline G_Z} ) + \alpha L_T( P^{G_X}, Q^{\overline G_Z} )$
                \\
                Update parameters: $W \leftarrow W - l_r \frac{ \partial {L} }{\partial W_{Enc}}$
                }
            }
            Get the final embedding result, $Z \leftarrow \phi(X, A|W)$
        }
    \label{alg:DMAGE}
    \end{algorithm}

}

%% file: Tab_info_4dataset.tex
\begin{table}[h]
    \centering
    \caption{Basis statistics of datasets}
    \begin{tabular}{@{}ccccc@{}}
    \toprule
    Dataset  & \#Nodes & \#Edges & \#Features & \#Classes \\ \midrule
    CORA     & 2,708  & 5,429  & 1,433     & 7        \\
    CiteSeer & 3,327  & 4,732  & 3,703     & 6        \\
    PubMed   & 19,717 & 44,338 & 500       & 3        \\
    Wiki     & 2,405  & 17,981 & 4,973     & 17       \\ \bottomrule
    \label{Tab_info_4dataset}
    \end{tabular}
\end{table}

%% file: Tab_result_node_clustering.tex
\clearpage

\begin{landscape}

    \begin{table*}[ht]
        \centering
        {
            \footnotesize
            \caption{
                Node clustering performance on PubMed, Cora, CiteSeer and Wiki. 
                Our DMAGE has the highest ACC (accuracy) scores in 11 of 12 metrixs.
                Concurrent Work are marked with \color{olivine}{olivine}.}
            \begin{tabular}{@{}cc|ccc|ccc|ccc|ccc@{}}
                \toprule
                \multirow{2}{*}{Method}                          & \multirow{2}{*}{Input}     & \multicolumn{3}{c|}{\pmb{CORA}}                & \multicolumn{3}{c|}{\pmb{CiteSeer}}          & \multicolumn{3}{c|}{\pmb{PubMed}}            & \multicolumn{3}{c}{\pmb{Wiki}}                      \\ 
                                                                &                            & ACC           & NMI           & F1            & ACC           & NMI           & F1            & ACC           & NMI           & F1           & ACC           & NMI           & F1            \\ \midrule
                k-means                                          & Feature                    & 34.7          & 16.7          & 25.4          & 38.5          & 17.0          & 30.5          & 57.3          & 29.1          & 57.4         & 33.4          & 30.2          & 24.5          \\
                Spectral-f                                       & Feature                    & 36.3          & 15.1          & 25.6          & 46.2          & 21.2          & 33.7          & 59.9          & 32.6          & 58.6         & 41.3          & 44.0          & 25.2          \\ \midrule
                Spectral-g                                       & Graph                      & 34.2          & 19.5          & 30.2          & 25.9          & 11.8          & 29.5          & 39.7          & 3.5           & 52.0         & 23.6          & 19.3          & 17.2          \\
            DeepWalk(2014) {\cite{perozzi_deepwalk_2014}}        & Graph                      & 46.7          & 31.8          & 38.1          & 36.2          & 9.7           & 26.7          & 61.9          & 16.7          & 47.1         & 38.5          & 32.4          & 25.7          \\
                DNGR(2016) {\cite{cao_deep_2016}}                & Graph                      & 49.2          & 37.3          & 37.3          & 32.6          & 18.0          & 44.2          & 45.4          & 15.4          & 17.9         & 37.6          & 35.9          & 25.4          \\ \midrule
                GAE(2016) {\cite{kipf_variational_2016}}         & Both                       & 53.3          & 40.7          & 42.0          & 41.3          & 18.3          & 29.1          & 64.1          & 23.0          & 49.3         & 17.3          & 11.9          & 15.4          \\
                VGAE(2016) {\cite{kipf_variational_2016}}        & Both                       & 56.0          & 38.5          & 41.5          & 44.4          & 22.7          & 31.9          & 65.5          & 25.1          & 51.0         & 28.7          & 30.3          & 20.5          \\
                MGAE(2017) {\cite{wang_mgae_2017}}               & Both                       & 63.4          & 45.6          & 38.0          & 63.6          & 39.8          & 39.5          & 43.9          & 8.2           & 42.0         & 50.1          & 48.0          & 39.2          \\
                ARGE(2018) {\cite{pan_adversarially_2018}}       & Both                       & 64.0          & 44.9          & 61.9          & 57.3          & 35.0          & 54.6          & 59.1          & 23.2          & 58.4         & 41.4          & 39.5          & 38.3          \\
                DANE(2018) {\cite{gao_deep_2018}}                & Both                       & 70.3          & 54.3          & 66.8          & 48.0          & 25.6          & 34.8          & 69.4          & 31.2          & 67.58        & 47.3          & 46.2          & 35.9          \\
                DGI (2018) {\cite{velickovic_deep_2018}}         & Both                       & 71.3          & 56.4          & 68.2          & 68.8          & 44.4          & 65.7          & 53.3          & 18.1          & 18.6         & -             & -             & -             \\
                AGC (2019) {\cite{zhang_attributed_2019}}        & Both                       & 68.9          & 53.7          & 65.6          & 67.0          & 41.1          & 62.5          & 69.8          & 31.6          & 68.7         & 47.7          & 45.2          & 35.7          \\
                DAEGC(2019) {\cite{wang_attributed_2019}}        & Both                       & 70.4          & 52.8          & 68.2          & 67.2          & 39.7          & 63.6          & 67.1          & 26.6          & 65.9         & -             & -             & -             \\
                ARGA(2020) {\cite{pan_learning_2020}}            & Both                       & 71.1          & 52.6          & 69.3          & 58.1          & 33.8          & 52.5          & 69.0          & 29.0          & 67.8         & -             & -             & -             \\
                AGE (2020) {\cite{cui_adaptive_2020}}            & Both                       & 71.2          & 55.9          & 68.2          & 56.9          & 34.8          & 54.4          & -             & -             & -            & 51.9          & 49.4          & 40.8          \\
    \color{olivine}{GIC (2020){ \cite{mavromatis_graph_2020}} }  & \color{olivine}Both        &\color{olivine} 72.5 & \color{olivine} 53.7    & \color{olivine} 69.4          &\color{olivine} 69.6          &\color{olivine} \pmb{45.3}    &\color{olivine} 65.4          &\color{olivine} 67.3          &\color{olivine} 31.9          &\color{olivine} 70.4         &\color{olivine} 50.5          &\color{olivine} 48.6          &\color{olivine} 43.8          \\ \midrule
                Ours(DMAGE-SED)                                  & Both                       & {74.2}        & {58.0}        & {69.8}        & {69.6}        & 44.1          & {66.3}        & {73.3}        & {35.8}        & {73.2}       & {52.3}        & {49.0}        & {46.8}       \\ 
                Ours(DMAGE-LOGI)                                 & Both                       & \pmb{74.3}    & {57.7}        & {69.7}        & {69.7}        & 43.8          & {66.3}        & \pmb{76.0}    & \pmb{36.8}    & \pmb{75.8}   & \pmb{53.8}    & \pmb{50.4}    & \pmb{47.6}   \\ 
                Ours(DMAGE-SED+LOGI)                             & Both                       & \pmb{74.3}    & \pmb{58.2}    & \pmb{70.0}    & \pmb{69.9}    & 43.5          & \pmb{66.5}    & {74.1}        & {36.0}        & {73.1}       & 52.1          & {50.0}        & {44.6}       \\ \bottomrule
                \end{tabular}
            \label{Tab_result_node_clustering}
        }
    \end{table*}
\end{landscape}

%% file: Tab_result_linkprediction.tex
\begin{landscape}
    \begin{table*}[htb]
    \centering
    {
        \footnotesize
    \caption{
        Link prediction performance on Cora and CiteSeer, and PubMed.
        Our DMAGE has the highest scores in Cora and CiteSeer.
        Concurrent Work are marked with \color{olivine}{olivine}.}
    \begin{tabular}{@{}c|cc|cc|cc@{}}
        \toprule
                                                                            & \multicolumn{2}{c|}{\pmb{CORA}}                                     & \multicolumn{2}{c|}{\pmb{CiteSeer}}                             & \multicolumn{2}{c}{\pmb{PubMed}}                  \\ 
                                                                            & AUC                       & AP                                & AUC                           & AP                        & AUC                        & AP                   \\ \midrule
        Spectral-g                                                          & 84.6 ($\pm$0.01)          & 88.5 ($\pm$0.01)                  & 80.5 ($\pm$0.01)              & 85.0 ($\pm$0.01)          & 84.2 ($\pm$0.02)           & 87.8 ($\pm$0.01)           \\
        DeepWalk(2014) {\cite{perozzi_deepwalk_2014}}                       & 83.1 ($\pm$0.01)          & 85.0 ($\pm$0.01)                  & 80.5 ($\pm$0.02)              & 83.6 ($\pm$0.01)          & 84.4 ($\pm$0.00)           & 84.1 ($\pm$0.01)           \\
        VGAE(2016) {\cite{kipf_variational_2016}}                           & 91.4 ($\pm$0.01)          & 92.6 ($\pm$0.01)                  & 90.8 ($\pm$0.02)              & 92.0 ($\pm$0.02)          & 96.4 ($\pm$0.00)           & 96.5 ($\pm$0.01)           \\
        DGI (2018) {\cite{velickovic_deep_2018}}                            & 89.8 ($\pm$0.8)           & 89.7 ($\pm$1.0)                   & 95.5 ($\pm$1.0)               & 95.7 ($\pm$1.0)           & 91.2 ($\pm$0.6)            & 92.2 ($\pm$0.5)            \\
        AGE (2020) {\cite{cui_adaptive_2020}}                               & 92.4 ($\pm$0.004)         & 93.2 ($\pm$0.003)                 & 92.4 ($\pm$0.003)             & 93.0 ($\pm$0.003)         & \pmb{96.8 ($\pm$0.001)} & \pmb{97.1 ($\pm$0.001)} \\
        \color{olivine} GIC (2020) { \cite{mavromatis_graph_2020}}          & \color{olivine}93.5 ($\pm$0.6)           & \color{olivine}93.3 ($\pm$0.7)                   & \color{olivine}97.0 ($\pm$0.5)               & \color{olivine}96.8 ($\pm$0.5)           & \color{olivine}93.7 ($\pm$0.3)            & \color{olivine}93.5 ($\pm$0.3)            \\ \midrule
        Ours(DMAGE-SED)                                                     & {96.0 ($\pm$0.3)}         & \pmb{96.8 ($\pm$0.2)}          & \pmb{98.17 ($\pm$0.15)}    & \pmb{98.32 ($\pm$0.11)}& 94.2 ($\pm$0.2)            & 93.8 ($\pm$0.4)            \\ 
        Ours(DMAGE-LOGI)                                                    & \pmb{96.7 ($\pm$0.2)}  & {96.0 ($\pm$0.2)}                 & {97.8 ($\pm$0.09)}            & {97.42 ($\pm$0.1)}        & 94.6 ($\pm$0.3)            & 94.7 ($\pm$0.2)            \\ 
        Ours(DMAGE-SED+LOGI)                                                & {95.4 ($\pm$0.1)}         & {96.0 ($\pm$0.3)}                 & 97.8 ($\pm$0.11)              & 97.92 ($\pm$0.08)         & 94.1 ($\pm$0.3)            & 95.2 ($\pm$0.4)            \\ \bottomrule
        \end{tabular}                       
        \label{Tab_result_linkprediction}
        }
    \end{table*}
\end{landscape}

%% file: Tab_result_ablation.tex
\begin{table}[tbp]
    \centering
        \caption{Ablation study result on node clustering ACC}
        \centering
        \begin{tabular}{@{}lcccc@{}}
            \toprule
                                            & \multicolumn{1}{l}{\pmb{CORA}} & \multicolumn{1}{l}{\pmb{CiteSeer}} & \multicolumn{1}{l}{\pmb{PubMed}} & \multicolumn{1}{l}{\pmb{Wiki}} \\ \midrule
                DMAGE                       & \pmb{74.15}                    & \pmb{69.7}                         & \pmb{76.35}                      & \pmb{51.51}           \\ 
                W/o augmentation            & 74.11                          & 69.61                              & 71.33                            & 48.64                    \\
                W/o FCA                     & 73.89                          & 68.37                              & 69.59                            & 50.14                    \\
                W/o ``soft" similarity      & 65.32                          & 66.78                              & 65.00                            & 43.45                    \\ \bottomrule
        \end{tabular}
        \label{Tab_result_ablation}
    \end{table}

%% file: appendix_nodeclustingmutirun.tex
\section{Statistical results of node clustering tasks}

Since most of the compared methods only report the highest results, we report the best results under 20 different random seeds in the main paper. The results of the mean and standard deviation are shown in Table.~\ref{appendixB_nodeclustingmutirun}.

\begin{table}[h]
    \centering
    \caption{Statistical results of node clustering tasks}
    \begin{tabular}{@{}cccc@{}}
        \toprule
                 & Acc(\%)              & NMI                   & F1                \\ \midrule
        CORA     & 73.56 ($\pm$0.78)    & 56.86 ($\pm$0.58)     & 69.38 ($\pm$0.96)     \\
        CiteSeer & 69.4 ($\pm$0.25)     & 43.46 ($\pm$0.27)     & 65.93 ($\pm$0.513)    \\
        PubMed   & 70.13 ($\pm$8.08)    & 31.94 ($\pm$5.60)     & 68.01 ($\pm$8.39)     \\ 
        Wiki     & 47.63 ($\pm$2.13)    & 48.86 ($\pm$0.99)     & 43.31 ($\pm$1.56)     \\ \bottomrule
        \end{tabular}
    \label{appendixB_nodeclustingmutirun}
    \end{table}

The values in parentheses are the standard deviations of 20 seeds. The experiments prove that our method has high stability in all three metrics. However, due to the excessive number of categories in the wiki dataset, the results have large fluctuations.

%% file: Tab_param.tex

\begin{table}[ht]
    \centering
    \caption{Hyperparameter settings for node clustering and link prediction. }
    \begin{tabular}{@{}cccc|cccc@{}}
    \toprule
                     & \#Nodes  & \#Edges   & \#Features & \#Classes   & $\nu_{latent}$       & $\alpha$    & $Q_p$  \\ \midrule
            CORA     & 2,708    & 5,429     & 1,433      & 7           & 0.001                & 1.0         & 50     \\
            CiteSeer & 3,327    & 4,732     & 3,703      & 6           & 0.003                & 0.5         & 80     \\
            PubMed   & 19,717   & 44,338    & 500        & 3           & 0.003                & 60          & 20     \\
            Wiki     & 2,405    & 17,981    & 4,973      & 17          & 0.02                 & 150         & 70     \\ \bottomrule
    \end{tabular}           
    \label{tab:Hyperparameter}
    \end{table}





%% file: appendix_visualization.tex
\begin{figure*}[!htb]
    \centering
    \includegraphics[width=4.5in]{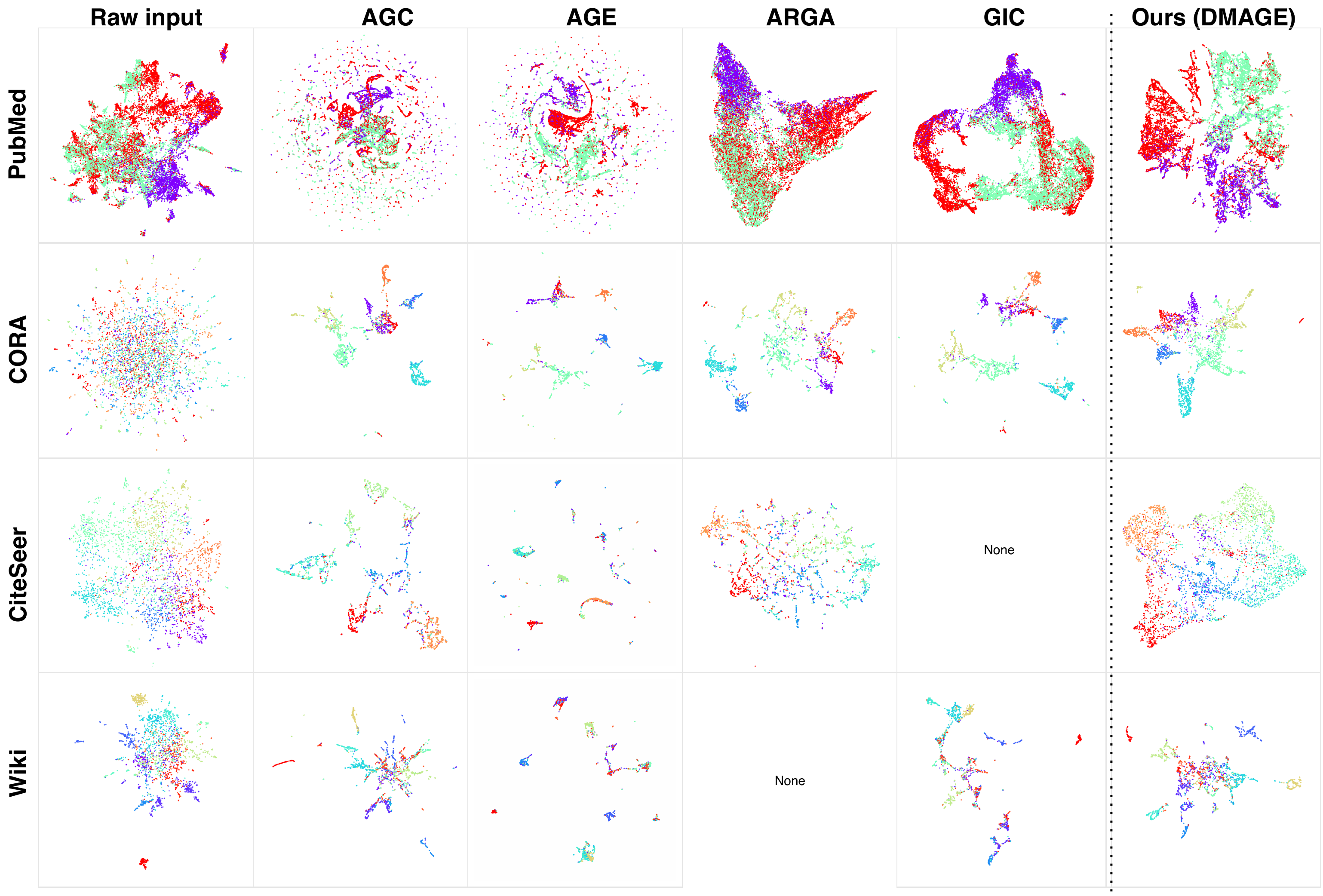}
      \caption{2D visualization of node representations on Pumbemd and Cora using UMAP. The different colors represent different classes. In the two displayed datasets, our method has less blending in the embedding result clusters, and all other methods have more color overlap. In addition, our method does not sever the connection between clusters while ensuring the subclustering, which indicates that the method retains richer information.}
      \label{fig:vis1_appendix}
  \end{figure*}